\documentclass[table]{article}

\usepackage{microtype}
\usepackage{graphicx}
\usepackage{booktabs} 

\usepackage{hyperref}

\usepackage[accepted]{icml2025}

\usepackage{amsmath}
\usepackage{amssymb}
\usepackage{mathtools}
\usepackage{amsthm}
\usepackage{graphicx}
\usepackage{url}
\usepackage{graphicx} 
\usepackage{booktabs}
\usepackage{tabularx}
\usepackage{pifont}
\usepackage{algorithmic}
\usepackage{textcomp}
\usepackage{multirow}
\usepackage{subcaption}
\usepackage{multicol}
\usepackage{xcolor}

\newcommand{\figref}[1]{Figure~\ref{#1}} 
\newcommand{\tabref}[1]{Table~\ref{#1}}
\newcommand{\secref}[1]{Section~\ref{#1}} 

\newcommand{\dilation}[0]{d}
\newcommand{\inputim}[0]{x}
\newcommand{\mask}[0]{m}
\newcommand{\prompt}[0]{p}
\newcommand{\segout}[0]{\hat{m}}
\newcommand{\dilateout}[0]{\hat{m}_d}
\newcommand{\refineout}[0]{\hat{m}_r}

\newcommand{\inversedilate}[0]{\hat{m}_d'}
\newcommand{\xhat}[0]{\hat{x}}
\newcommand{\restoreout}[0]{\dot{x}}
\newcommand{\SegModel}[0]{S}
\newcommand{\RefineModel}[0]{R}

\usepackage[capitalize,noabbrev]{cleveref}

\theoremstyle{plain}

\theoremstyle{definition}

\theoremstyle{remark}

\usepackage[textsize=tiny]{todonotes}

\icmltitlerunning{Blind Visible Watermark Removal with Morphological Dilation}

\begin{document}

\twocolumn[
\icmltitle{Blind Visible Watermark Removal with Morphological Dilation}



\icmlsetsymbol{equal}{*}

\begin{icmlauthorlist}
\icmlauthor{Preston K. Robinette}{sch}
\icmlauthor{Taylor T. Johnson}{sch}
\end{icmlauthorlist}

\icmlaffiliation{sch}{Department of Computer Science, Vanderbilt University, Nashville TN, USA}

\icmlcorrespondingauthor{Preston K. Robinette}{preston.k.robinette@vanderbilt.edu}
\icmlcorrespondingauthor{Taylor T. Johnson}{taylor.johnson@vanderbilt.eud}

\icmlkeywords{Security, Watermarking, Information Hiding}

\vskip 0.3in
]



\printAffiliationsAndNotice{} 


\begin{abstract}
Visible watermarks pose significant challenges for image restoration techniques, especially when the target background is unknown. Toward this end, we present MorphoMod, a novel method for automated visible watermark removal that operates in a blind setting---without requiring target images. Unlike existing methods, MorphoMod effectively removes opaque and transparent watermarks while preserving semantic content, making it well-suited for real-world applications. Evaluations on benchmark datasets, including the Colored Large-scale Watermark Dataset (CLWD), LOGO-series, and the newly introduced Alpha1 datasets, demonstrate that MorphoMod achieves up to a 50.8\% improvement in watermark removal effectiveness compared to state-of-the-art methods. Ablation studies highlight the impact of prompts used for inpainting, pre-removal filling strategies, and inpainting model performance on watermark removal. Additionally, a case study on steganographic disorientation reveals broader applications for watermark removal in disrupting high-level hidden messages. MorphoMod offers a robust, adaptable solution for watermark removal and opens avenues for further advancements in image restoration and adversarial manipulation.
\end{abstract}

\section{Introduction}
\label{sec:intro}
Watermarks are widely employed as a digital rights management (DRM) tool to protect intellectual property in images, videos, and other multimedia content~\cite{cox2002digital,singh2013survey}. Invisible watermarks, imperceptible to the human eye, rely on sophisticated signal-processing techniques to encode information directly into the digital media. Visible watermarks, on the other hand, are prominently displayed overlays that serve as a clear visual deterrent against unauthorized distribution or misuse.

Visible watermarks are often used due to their immediate perceptibility and typically take the form of logos, text, or patterns strategically placed over an image. Despite their effectiveness, visible watermarks are not impervious to attacks. Techniques such as traditional inpainting~\cite{huang2004attacking,xu2017automatic,qin2018visible} and  Independent Component Analysis~\cite{pei2006novel} can be employed to remove or obscure these watermarks. The rise of generative models and diffusion-based inpainting methods has further amplified this vulnerability, enabling highly realistic reconstruction of the original image with minimal artifacts~\cite{cheng2018large,hertz2019blind,cun2021split,liang2021visible, liu2021wdnet, sun2023denet}. Addressing these challenges requires a deeper understanding of both watermarking techniques and the methods employed to remove them.
\begin{figure}[t]
    \centering
    \includegraphics[width=\linewidth]{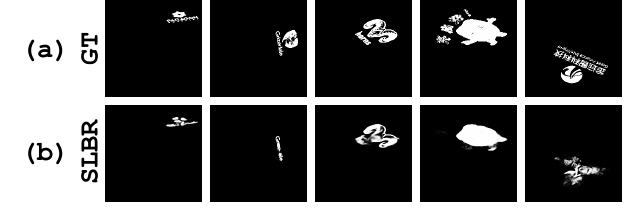}
    \caption{Example watermark localization masks generated with SLBR~\cite{liang2021visible}, a visible watermark removal method, during the removal process. The predicted masks are conservative compared to the ground truth (GT), leading to watermark residuals in the processed image.}
    \label{fig:mask_ex}
\end{figure}
\begin{figure*}[t]
    \centering
    \includegraphics[width=\textwidth]{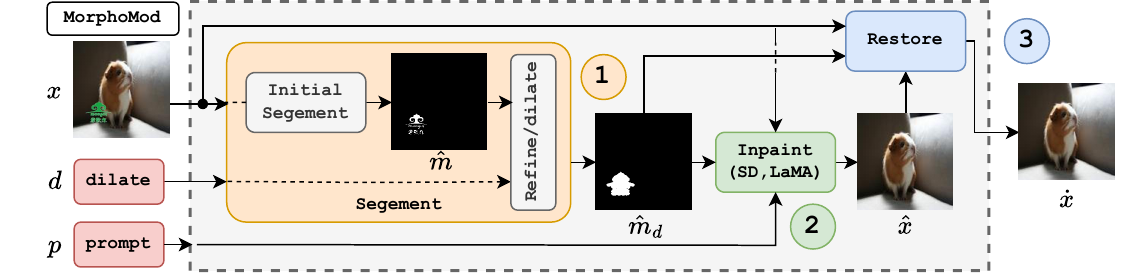}
    \caption{A diagram of MorphoMod---the proposed automated visible watermark removal method with mask refinement and generative inpainting. MorphoMod consists of three main phases: (1) \textit{segment}, (2) \textit{inpaint}, and (3) \textit{restore}. In \textit{segment}, an input image $\inputim$ and dilation parameter $\dilation$ are used to produce a refined and dilated mask $\dilateout$. The input image, a prompt $\prompt$, and the refined mask are then used to \textit{inpaint} the detected region resulting in a cleaned image $\xhat$. In \textit{restore}, this cleaned image, the input image, and the refined mask are used to generate the final restored image with the removed watermark $\restoreout$.}
    \label{fig:morphomod}
\end{figure*}
\begin{figure}[t]
    \centering
    \includegraphics[width=\linewidth]{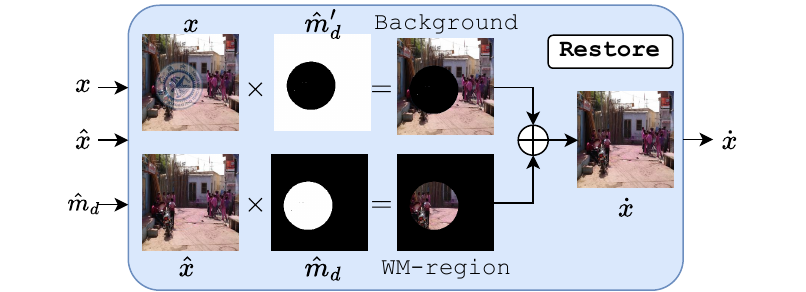}
    \caption{An example restoration. A generated image is combined with the refined mask to select only the watermarked region. The input image and the inverse of the refined mask are used to select the background region. The watermarked region and background region are then combined to create the restored image.}
    \label{fig:restore_ex}
\end{figure}

As such, we  develop a novel automated pipeline stemming from three key observations: \textit{(1) Two-Stage Pipelines$\rightarrow$} the majority of deep learning methods follow a two-stage pipeline, beginning with the identification of a watermark mask followed by image restoration,
\textit{(2) Conservative Mask Predictions$\rightarrow$} the masks generated by current methods are often smaller than the true watermark (see \figref{fig:mask_ex}) leading to incomplete removal, and \textit{(3) Dependency on Known Backgrounds$\rightarrow$} all current deep-learning approaches require prior knowledge of the target background during the training process (\textit{non-blind}).

Building upon these key observations, we propose a \textit{blind} visible watermark removal framework that employs morphological dilation in conjunction with inpainting called \textit{MorphoMod}. By eliminating the need for prior knowledge of the background and focusing on robust mask generation, our method mimics real-world scenarios. The contributions of this work are the following: \textbf{(1) Development of MorphoMod$\rightarrow$} we introduce a novel, blind watermark removal method effective for real-world scenarios involving datasets without known targets, \textbf{(2) Introduction of novel datasets$\rightarrow$} we propose two new datasets, Alpha1-S and Alpha1-L, featuring non-transparent, fully opaque watermarks, designed to better reflect real-world watermarking challenges, \textbf{(3) Introduction of novel evaluation metrics$\rightarrow$} we propose new metrics specifically tailored to assess watermark removal performance in the absence of known target images, \textbf{(4) Comprehensive experimental analysis$\rightarrow$} we perform extensive ablation studies to evaluate the impact of various factors, including inpainting models, prompt design, pre-removal filling strategies on watermark removal and image quality, \textbf{(5) Case study on steganographic disorientation$\rightarrow$} we introduce a novel toy problem and dataset to showcase MorphoMod's potential in disrupting visible steganographic messages.

\section{Related Works}
\label{sec:related_works}
In this section, we present research related to inpainting and deep learning-based visible watermark removal (VWR). We also provide key observations from deep learning based VWR.

\textbf{Inpainting.} Inpainting aims to restore missing or corrupted regions in an image. Traditional methods often rely on partial differential equations (PDEs)~\cite{bertalmio2006pde,schonlieb2015partial} or patch synthesis~\cite{criminisi2003object} while deep learning approaches~\cite{pathak2016context,nazeri2019edgeconnect,cai2022image} leverage large datasets to learn complex features, often employing generative adversarial networks (GANs). Recent advancements harness denoising diffusion probabilistic models (DDPMs)~\cite{ho2020denoising} for inpainting by iteratively refining noisy versions of an image. Methods like Stable Diffusion~\cite{rombach2021highresolution} and LaMa~\cite{suvorov2021resolution} achieve high-quality results.

\textbf{Deep Learning Watermark Removal Methods.} Recent advancements in VWR have been driven by deep learning-based approaches. In~\cite{cheng2018large}, the authors introduce a seminal image-to-image translation approach driven by the object detection of watermarks which are then refined using a convolution neural network (CNN). To bypass the reliance on detection-based components, \cite{li2019towards} and \cite{cao2019generative} proposed generative adversarial network (GAN) techniques. In~\cite{hertz2019blind}, the authors introduce a multi-task learning framework utilizing a single encoder with multi-decoder architecture to reconstruct the background, motif (watermark) mask, and motif (watermark) image. \cite{cun2021split} introduce a two-stage network for multi-task decoding and then refinement called \textit{SplitNet}. \cite{liang2021visible} refine mask predictions even further using a coarse-to-fine strategy known as \textit{SLBR}. Instead of estimating only the mask of the watermark, \textit{WDNet}~\cite{liu2021wdnet} attempts to predict the mask, opacity and color of the watermark. \textit{DENet} attempts to disentangle watermark and image embeddings in the feature space, employing contrastive learning to ensure task-specific representation for watermark removal and background reconstruction~\cite{sun2023denet}. In ~\cite{leng2024removing}, the authors also implement a two-stage approach consisting of watermark localization and background content restoration, known as \textit{RIRCI}.

\textbf{Key Observations.} \textbf{[1-2]} Most deep learning methods for watermark removal use a mask prediction component to guide background restoration as shown in \figref{fig:restore_ex}. This procedure boosts image quality metrics by enabling a focus on reconstructing or refining the watermarked area. However, the accuracy of the predicted mask is crucial, as it directly impacts the final image composition. Many prior methods focus on improving the reconstructed image while neglecting mask refinement. \figref{fig:mask_ex} illustrates example predicted masks for SLBR, where under-prediction often leaves watermark residuals in the sanitized image. \textbf{[3]} Another key limitation is the reliance on access to watermark-free images during training. Training datasets typically include \textit{watermarked images}, \textit{watermark masks}, and \textit{watermark-free images}~\cite{liu2021wdnet,cun2021split,cheng2018large}. In real-world scenarios, matching the distribution of watermarked and watermark-free images is highly unlikely. For instance, when watermarked images are scraped from platforms like Shutterstock.com, the true background behind the watermark is unknown. This raises a critical challenge: how can these methods be effectively evaluated when ground truth watermark-free images are unavailable in practice?

\section{MorphoMod}
Based on the key observations from \secref{sec:related_works}, our work introduces a blind watermark removal method that combines morphological dilation with diffusion-based inpainting to address the limitations of existing methods. Unlike traditional two-stage pipelines, which focus on the refinement of the image, our approach integrates mask prediction and mask refinement into a unified framework, ensuring that generated masks more closely align with the true watermark extent. By eliminating the need for background priors, our method offers a practical and scalable solution for watermark removal in real-world applications. MorphoMod consists of three main stages: 1) \textit{segment}, 2) \textit{inpaint}, and 3) \textit{restore}, as shown in \figref{fig:morphomod}. We discuss each stage in more detail below.

\subsection{Segment}
\textbf{Initial Segmentation.} The segmentation stage consists of two components: (1) initial segmentation and (2) refinement. For the initial segmentation, we use the best-performing semantic segmentation model from SLBR, SplitNet, WDNet, DENet, or RIRCI (which relies on SLBR for mask prediction) for each dataset. Let $\SegModel$~be the segmentation model, $\inputim \in \mathcal{R}^{3xHxW}$ the watermarked image, and $\segout \in \mathcal{R}^{1xHxW}$ the generated segmentation mask. The initial segmentation produces an initial mask $\segout$ s.t. $\SegModel(\inputim) = \segout$.

\textbf{Refinement.} The refinement step uses a U-Net model and morphological dilation to improve the initial mask by addressing underprediction. The U-Net Refine Model $\RefineModel$ refines the mask, taking $\inputim$ and $\segout$ as input and producing $\refineout$ such that $\RefineModel(\inputim, \segout) = \refineout$. The model is trained with the following loss:%
\begin{equation}
\mathcal{L}_{\text{total}} = \mathcal{L}_{\text{Dice}}(\refineout, \mask ) + \mathcal{L}_{\text{BCE}}(\refineout, \mask)
\end{equation}%
Here, \(\mathcal{L}_{\text{Dice}}\) measures overlap between $\refineout$ and the ground truth mask $\mask$, while \(\mathcal{L}_{\text{BCE}}\) evaluates pixel-wise differences, promoting both global and local accuracy. Morphological dilation is then applied to expand mask boundaries, ensuring all residual watermark artifacts are captured. Dilation is defined as:
\begin{equation}
\dilateout = \max_{s \in S} \big(\refineout(s) \oplus \mathbf{K}(s)\big)
\end{equation}
where \(\mathbf{K}(s)\) is a structuring element (e.g., disk or square) and \(\oplus\) the dilation operation. Larger kernels (e.g., \(7 \times 7\)) expand boundaries significantly, while smaller kernels (e.g., \(3 \times 3\)) provide finer adjustments. The kernel size is set by a dilation parameter $\dilation$.

\textbf{Overview.} The segment phase takes as input a watermarked image $\inputim$ and a dilation parameter $\dilation$ to produced a refined mask $\dilateout$ for inpainting, i.e., $\mathrm{Segment}(\inputim, \dilation) = \dilateout$.  
\begin{figure*}[tbh!]
\vskip 0.2in
\begin{center}
\begin{subfigure}{\textwidth}
    \centerline{\includegraphics[width=\textwidth]{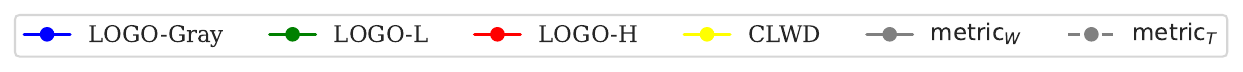}}
\end{subfigure}
\begin{subfigure}{0.33\textwidth}
    \centerline{\includegraphics[width=\columnwidth]{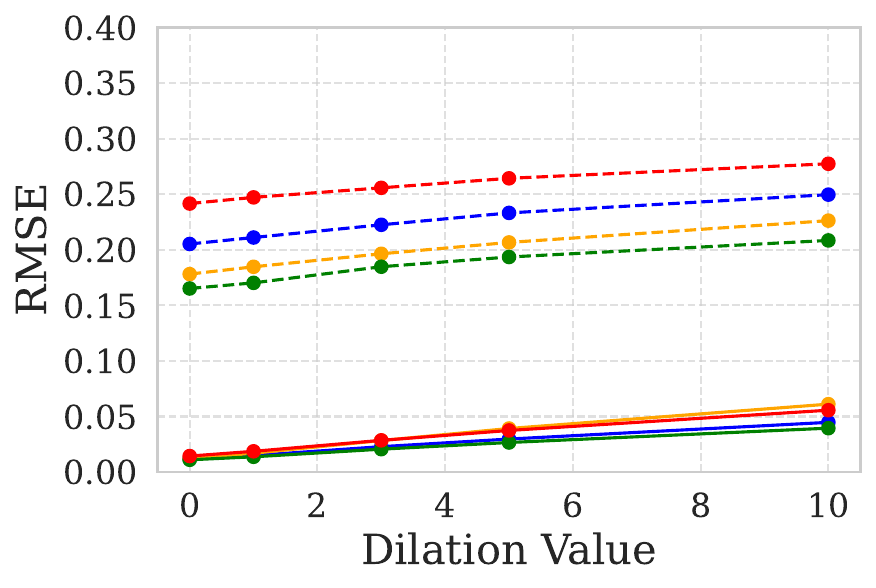}}
    \caption{RMSE}
    \label{fig:dilate_rmse}
\end{subfigure}
\begin{subfigure}{0.33\textwidth}
    \centerline{\includegraphics[width=\columnwidth]{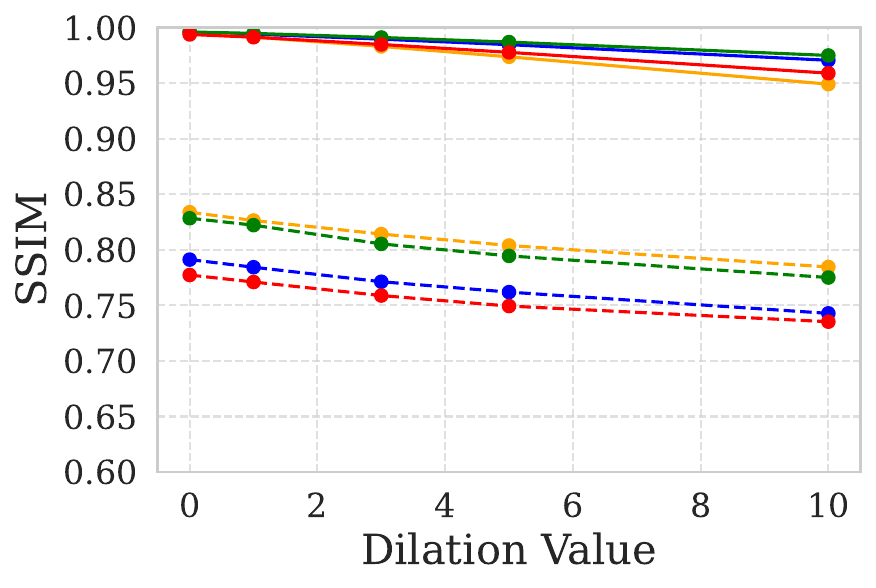}}
    \caption{SSIM}
    \label{fig:dilate_ssim}
\end{subfigure}
\begin{subfigure}{0.33\textwidth}
    \centerline{\includegraphics[width=\columnwidth]{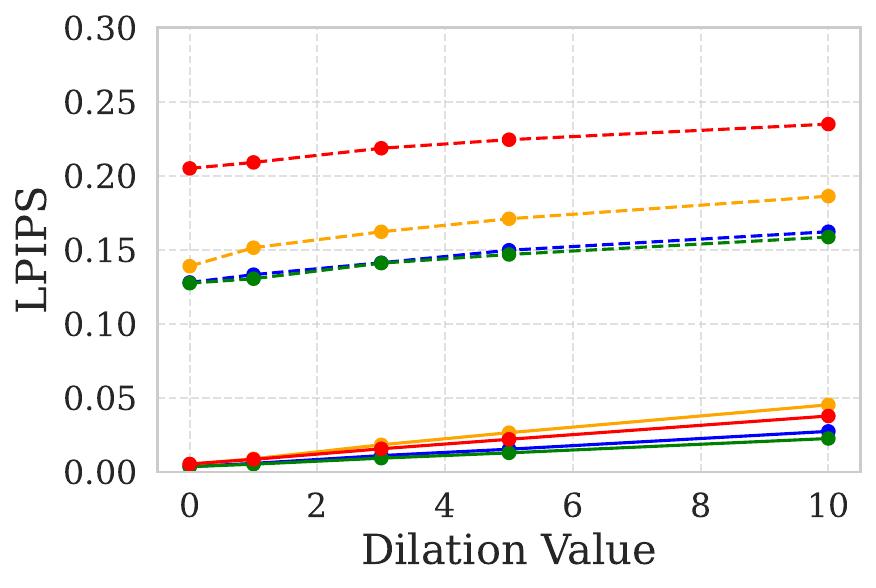}}
    \caption{LPIPS}
    \label{fig:dilate_lpips}
\end{subfigure}
\caption{Watermark removal (WR) and semantic preservation (SP) metrics for MorphoMod on the CLWD and LOGO-series datasets across various dilation values $\dilation$.}
\label{fig:dilate}
\end{center}
\vskip -0.2in
\end{figure*}

\subsection{Inpaint}
The refined mask $\dilateout$ is then combined with the watermarked image $\inputim$ and a prompt $\prompt$ for the inpaint phase. For the inpaint model, we make use of publicly available denoising diffusion probabilistic models (DDPM) like Stable Diffusion and LaMa. The output of the inpaint phase is a cleaned image $\xhat$, i.e., $\mathrm{Inpaint}(\inputim, \dilateout, \prompt) = \xhat$, where $\xhat \in \mathcal{R}^{3xHxW}$.

\subsection{Restore}
In the restore phase, the background from the watermarked image is then combined with the watermark region of the cleaned image. An example of this is shown in \figref{fig:restore_ex}. Here, the background region is created by multiplying the original image $\inputim$ with the inverse of the provided mask $\inversedilate$, s.t. $\inversedilate = 1 - \dilateout$. The cleaned watermark region is created by multiplying the processed output $\xhat$  with the refined mask $\dilateout$. The background is then combined with the watermark region to produce the restored image $\restoreout$, i.e., $\mathrm{Restore}(\inputim, \dilateout, \xhat) = \restoreout$.

\subsection{Blind Metrics} 
\label{sec:metrics}
In this work, we introduce novel metrics to evaluate watermark removal under real-world conditions where true watermark-free targets are unavailable. These metrics estimate both the effectiveness of watermark removal (WR) and the semantic preservation (SP) of the original image. We base these blind metrics on common VWR metrics: root mean squared error ($\mathrm{RMSE}$), structural similarity index measurement ($\mathrm{SSIM}$), and Learned Perceptual Image Patch Similarity ($\mathrm{LPIPS}$)~\cite{cheng2018large,hertz2019blind,cun2021split,liang2021visible, liu2021wdnet, sun2023denet}. \textbf{Watermark Removal.} Watermark removal is assessed by comparing the watermarked region in the original image to the corresponding region in the output image, using the ground truth mask. Metrics include $\mathrm{RMSE}_W$, $\mathrm{SSIM}_W$, and $\mathrm{LPIPS}_W$, where high $\mathrm{RMSE}_W$, low $\mathrm{SSIM}_W$, and high $\mathrm{LPIPS}_W$ indicate successful watermark removal. \textbf{Semantic Preservation.} Semantic preservation evaluates the background by comparing the background region of the watermarked image with that of the output image. The background region is determined by: $\text{background} = \inputim * (1-\mask)$, where $\inputim$~is the watermarked image and $\mask$~is the ground truth mask. Metrics $\mathrm{RMSE}_T$, $\mathrm{SSIM}_T$, and $\mathrm{LPIPS}_T$ are used, with low $\mathrm{RMSE}_T$, high $\mathrm{SSIM}_T$, and low $\mathrm{LPIPS}_T$ indicating effective background preservation.

\textbf{Tradeoff.} Balancing watermark removal and semantic preservation is critical, as optimizing one often compromises the other. A smaller mask may preserve semantics but perform poorly in watermark removal, while a larger mask improves removal at the cost of background integrity. In this work, we prioritize watermark removal.

\begin{table*}[tbh!]
\caption{Watermark removal (WR) and semantic preservation (SP) metrics for various visible watermark removal methods on the Alpha1-S and Alpha1-L datasets. The \colorbox{green!30}{green} blocks and \colorbox{red!30}{red} represent the best and worst in each category respectively.}
\label{tab:alpha_expr}
\begin{center}
\begin{small}
\begin{sc}
\begin{tabular}{p{0.15cm}|c|ccc|ccc}
\toprule
 & \multirow{2}{*}{Model} &  \multicolumn{3}{c|}{Alpha1-S} & \multicolumn{3}{c}{Alpha1-L} \\ 
 & & $RMSE_W (\uparrow)$ & $SSIM_W (\downarrow)$ & $LPIPS_W (\uparrow)$ & $RMSE_W (\uparrow)$ & $SSIM_W (\downarrow)$ & $LPIPS_W (\uparrow)$ \\
 %
%
\cline{2-8}
\multirow{7}{*}{\rotatebox[origin=c]{90}{\textbf{WR}}} & SLBR  & 0.0786 & 0.9189 & 0.0691  & 0.0552  & 0.9478 & 0.0480  \\ 
& SplitNet  & \cellcolor{red!30}0.0044 & \cellcolor{red!30}0.9993 & \cellcolor{red!30}0.0003  & \cellcolor{red!30}0.0032  & \cellcolor{red!30}0.9994 & \cellcolor{red!30}0.0004  \\ 
& WDNet  & 0.0676 & 0.9407 & 0.0355  & 0.0606 & 0.9633 & 0.0338  \\ 
& DENet-g  & 0.0841 & 0.9220 & 0.0479  & 0.0381 & 0.9848 & 0.0118  \\ 
& DENet-h  & 0.0660 & 0.9482 & 0.0388  & 0.0875 & 0.9352 & 0.0636  \\ 
& DENet-l  & 0.0195 & 0.9879 & 0.0063  & 0.0038 & 0.9993 & 0.0005  \\ 
& $\text{morphomod}_3$*  & \cellcolor{green!30}0.1165 & \cellcolor{green!30}0.8855 & \cellcolor{green!30}0.0763  & \cellcolor{green!30}0.1778 & \cellcolor{green!30}0.8523 & \cellcolor{green!30}0.1180  \\
\cline{2-8}
& \% Improve & 27.8\% & 4.1\% & 37.2\% & 50.8\% & 9.7\% & 46.1\% \\
%
%
\bottomrule
 & Model & $RMSE_T (\downarrow)$ & $SSIM_T (\uparrow)$ & $LPIPS_T (\downarrow)$ & $RMSE_T (\downarrow)$ & $SSIM_T (\uparrow)$ & $LPIPS_T (\downarrow)$ \\
\cline{2-8}
\multirow{7}{*}{\rotatebox[origin=c]{90}{\textbf{SP}}} & SLBR  & 0.0067 & 0.9980 & 0.0036  & 0.0077 & 0.9984 & 0.0024  \\ 
& SplitNet  & \cellcolor{green!30}0.0041 & \cellcolor{green!30}0.9993 & \cellcolor{green!30}0.0012  & 0.0049 & 0.9993 & 0.0012  \\ 
& WDNet  & \cellcolor{red!30}0.0212 & \cellcolor{red!30}0.9845 & \cellcolor{red!30}0.0180  & \cellcolor{red!30}0.0255 & \cellcolor{red!30}0.9869 & \cellcolor{red!30}0.0134  \\ 
& DENet-g  & 0.0056 & 0.9985 & 0.0026  & 0.0049 & 0.9990 & 0.0016  \\ 
& DENet-h  & 0.0075 & 0.9985 & 0.0025  & 0.0176 & 0.9954 & 0.0070  \\ 
& DENet-l  & 0.0043 & \cellcolor{green!30}0.9993 & \cellcolor{green!30}0.0012  & \cellcolor{green!30}0.0027 & \cellcolor{green!30}0.9997 & \cellcolor{green!30}0.0004  \\ 
& $\text{morphomod}_3$* & 0.0157 & 0.9934 & 0.0060  & 0.0183 & 0.9928 & 0.0057  \\ 
\bottomrule
\end{tabular}
\end{sc}
\end{small}
\end{center}
\end{table*}

\begin{figure*}[t]
    \centering
    \includegraphics[width=0.90\linewidth]{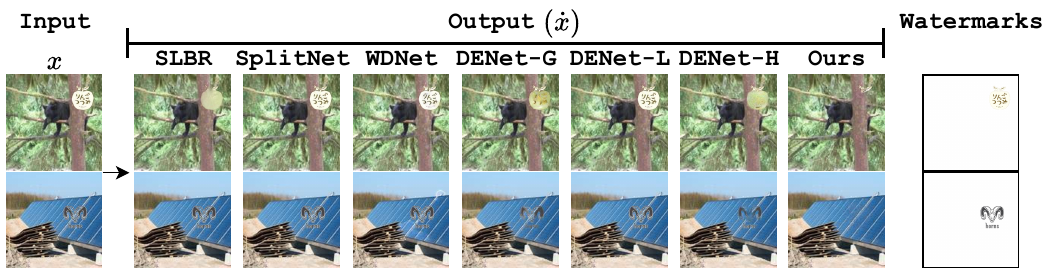}
    \caption{Visible watermark removal results on the Alpha1-S dataset. Our method, \textit{MorphoMod}, is the only method to successfully remove the opaque watermarks.}
    \label{fig:alpha_images}
\end{figure*}


\section{Experiments}
In this section, we introduce the datasets, implementation details, and results of two experiments used to evaluate MorphoMod. 

\subsection{Datasets and Implementation Details}
\textbf{Experiment 1.} \textit{Baseline performance and effect of the dilation parameter $\dilation$.} To determine the baseline performance of MorphoMod, we evaluate on four common watermarking datasets: 1) Colored Large-scale Watermark Dataset (CLWD)~\cite{liu2021wdnet}, 2) LOGO-Gray, 3) LOGO-L, and 4) LOGO-H~\cite{cun2021split}. Please see the Appendix for more details on these datasets.  Using these datasets, we then evaluate MorphoMod with dilation values $d \in \{0, 1, 3, 5, 10\}$ and prompt $p = \text{``Remove.''}$ and record watermark removal and semantic preservation metrics as described in \secref{sec:metrics}. 

\textbf{Experiment 2.} \textit{Comparison against existing methods on two new datasets: Alpha1-S and Alpha1-L.} Previous methods rely on datasets with transparent watermarks and known target images, which do not reflect all real-world scenarios where watermarks could be opaque and targets are unknown. To address this, we introduce Alpha1-S and Alpha1-L, featuring non-transparent watermarks with no assumed targets.

\textbf{Alpha1 Datasets:} \textbf{Alpha1-S}mall and \textbf{Alpha1-L}arge contain 12K training and 2K testing samples from ImageNet~\cite{deng2009imagenet} with opaque watermarks from the CLWD dataset. Each sample consists of a watermarked image and a corresponding mask. The watermarks cover 6\% of the image area in Alpha1-S and 35\% in Alpha1-L.

Using these datasets, we evaluate MorphoMod ($\dilation=3$, $p=\text{``Remove.''}$) against SLBR~\cite{liang2021visible}, SplitNet~\cite{cun2021split}, WDNet~\cite{liu2021wdnet}, and DENet variants trained on LOGO datasets~\cite{sun2023denet}. These models, which rely on target images during training, were pretrained on CLWD or LOGO datasets and evaluated on Alpha1-S and Alpha1-L using watermark removal and semantic preservation metrics (\secref{sec:metrics}). This setup mimics transfer learning by assessing how well pretrained models perform on out-of-distribution, web-scraped data.%

\begin{table*}[t]
\caption{Inpainting experimental results of various experiments on the Alpha1-S and CLWD datasets for $\mathrm{RMSE}_W$ ($R_W$), $\mathrm{SSIM}_W$ ($S_W$), $\mathrm{LPIPS}_W$ ($LP_W$), BRISQUE (B), NIQE (N), and PIQE (P).  }
\label{tab:comb_expr}
\vskip 0.1in
\begin{center}
\begin{small}
\begin{sc}
\begin{tabular}{p{0.1cm}|p{1.1cm}|cccp{0.70cm}p{0.70cm}p{0.70cm}|cccp{0.70cm}p{0.70cm}p{0.70cm}}
\toprule
 & \multirow{2}{*}{Var} &  \multicolumn{6}{c|}{Alpha1-S} & \multicolumn{6}{c}{CLWD} \\ 
 \cline{3-14}
&  & $R_W (\uparrow)$ & $S_W (\downarrow)$ & $LP_W (\uparrow)$ & B $(\downarrow)$ & N$(\downarrow)$ & P$(\downarrow)$ & $R_W (\uparrow)$ & $S_W (\downarrow)$ & $LP_W (\uparrow)$ & B$(\downarrow)$ & N$(\downarrow)$ & P  $(\downarrow)$\\
 \midrule
\multirow{5}{*}{\rotatebox[origin=c]{90}{\textbf{5.2}}}& P1 & 0.369 & 0.599 & 0.317 &  38.88 & 4.96 & 43.49 & 0.216 & 0.791 & 0.176 &  25.24 & 4.39 & 38.49 \\
 & P2 & 0.369 & 0.599 & 0.318 &  39.34 & 4.97 & 43.29 & 0.211 & 0.795 & 0.173 &  25.41 & 4.37 & 38.47 \\
 & P3 & 0.373 & 0.598 & 0.317 &  38.96 & 4.96 & 43.23 & 0.223 & 0.786 & 0.181 &  24.78 & 4.38 & 38.45 \\
 & P4 & 0.370 & 0.599 & 0.317 &  39.21 & 4.95 & 43.34 & 0.214 & 0.793 & 0.174 &  25.45 & 4.34 & 38.41 \\
 & P5 & 0.371 & 0.598 & 0.317 &  39.18 & 4.99 & 43.33 & 0.216 & 0.792 & 0.176 &  24.86 & 4.36 & 38.28 \\
 \midrule
\multirow{5}{*}{\rotatebox[origin=c]{90}{\textbf{5.3}}} & Normal & 0.369 & 0.599 & 0.317 &  39.82 & 4.95 & 44.08 & 0.214 & 0.794 & 0.174 &  25.97 & 4.30 & 38.57 \\
&  BG & 0.369 & 0.600 & 0.316 &  39.84 & 4.94 & 44.13 & 0.213 & 0.795 & 0.174 &  24.92 & 4.25 & 37.22 \\
&  Black & 0.370 & 0.600 & 0.316 &  39.84 & 4.94 & 44.10 & 0.213 & 0.795 & 0.175 &  25.04 & 4.26 & 37.36 \\
&  Gray & 0.370 & 0.599 & 0.316 &  39.84 & 4.93 & 44.14 & 0.213 & 0.794 & 0.174 &  24.95 & 4.28 & 37.36 \\
&  White & 0.371 & 0.598 & 0.316 &  39.83 & 4.93 & 44.12 & 0.213 & 0.794 & 0.174 &  24.82 & 4.27 & 37.28 \\
 \midrule
\multirow{3}{*}{\rotatebox[origin=c]{90}{\textbf{5.4}}} & SD2 &  0.369 & 0.599 & 0.317 &  39.12 & 4.94 & 43.26 & 0.214 & 0.794 & 0.174 &  25.97 & 4.30 & 38.57 \\
&  SDXL & 0.334 & 0.625 & 0.299 & 39.63 & 4.91 & 44.06 & 0.198 & 0.800 & 0.169 & 24.49 & 4.36 & 37.91 \\
&  LaMa & 0.350 & 0.594 & 0.247 & 39.59 & 4.92 & 43.37 & 0.167 & 0.837 & 0.111 & 24.18 & 4.35 & 36.97 \\
 
\bottomrule
\end{tabular}
\end{sc}
\end{small}
\end{center}
\vskip -0.1in
\end{table*} 
\subsection{Results}
\textbf{Experiment 1.} Figure~\ref{fig:dilate} presents the results of Experiment~1, where the solid lines represent the watermark removal metrics and the dashed lines represent the semantic preservation metrics. For each dataset and dilation value, the high $\mathrm{RMSE}_W$, low $\mathrm{SSIM}_W$, and high $\mathrm{LPIPS}_W$ values indicate effective watermark removal. As the dilation value increases, more of the actual watermark is targeted, leading to stronger removal performance.

However, this improvement also results in a greater loss of the original image content, as shown by the trends in the semantic preservation metrics. This underscores the inherent trade-off between thorough watermark removal and preserving the image’s semantic integrity. Notably, MorphoMod removes the watermark across all dilation values tested without excessively compromising semantic quality, confirming its effectiveness for visible watermark removal.

\textbf{Experiment 2.} The results for Experiment~2 are shown in \tabref{tab:alpha_expr}. In terms of watermark removal, MorphoMod achieves the strongest performance across all metrics, as indicated by the high \(\mathrm{RMSE}_W\), low \(\mathrm{SSIM}_W\), and high \(\mathrm{LPIPS}_W\). For the Alpha1-S and Alpha1-T datasets, the best \textit{target-trained} models are DENet-G and SLBR, respectively. This is highlighted by the image results in \figref{fig:alpha_images}. In contrast, for semantic preservation, SplitNet and DENet-L deliver the highest performance, as shown by their low \(\mathrm{RMSE}_W\), high \(\mathrm{SSIM}_W\), and low \(\mathrm{LPIPS}_W\). This is expected given that better retention of the original background boosts semantic preservation metrics. If the mask is smaller than the true watermark area, more of the original background remains in the restored image, thus raising the semantic metrics but leaving behind noticeable watermark artifacts.

\begin{table}[t]
\caption{Segmentation performance (IoU and F1 Score) for various SAM schemes. }
\label{tab:eval_sam}
\vskip 0.1in
\begin{center}
\begin{small}
\begin{sc}
\begin{tabular}{c|cc|cc}
\toprule
 \multirow{2}{*}{SAM2 Method} &  \multicolumn{2}{c|}{Alpha1-S} & \multicolumn{2}{c}{CLWD} \\ 
 \cline{2-5}
  & IoU & F1 Score & IoU & F1 Score \\
 \midrule
 Auto Mask & 0.327 & 0.419 & 0.155 & 0.213 \\
 GT BBox & \cellcolor{green!30}0.543 & \cellcolor{green!30}0.677 & 0.524 & 0.653 \\
 GT Points & 0.064 & 0.091 & 0.112 & 0.161 \\
 GDINO BBox & 0.391 & 0.496 & 0.290 & 0.370 \\
 SLBR & 0.203 & 0.245 & \cellcolor{green!30}0.745 &  \cellcolor{green!30}0.825\\
 
\bottomrule
\end{tabular}
\end{sc}
\end{small}
\end{center}
\vskip -0.1in
\end{table} 
\begin{table}[t]
\caption{Prompts used for inpainting.}
\label{tab:prompts_small}
\vskip 0.1in
\begin{center}
\begin{small}
\begin{tabular}{c|l}
    \toprule
        \textsc{ID} & \textsc{ Prompt}  \\
        \midrule
        \textsc{P1} & Remove. \\
        \textsc{P2} & Fill in the background. \\
        \textsc{P3} & Erase the mark and restore the original. \\
        \textsc{P4} & Blend into the surrounding area. \\
        \textsc{P5} & Reconstruct the missing details. \\
    \bottomrule
\end{tabular}
\end{small}
\end{center}
\vskip -0.1in
\end{table} %

\textbf{Takeaways.} \textit{Overall, MorphoMod proves highly effective for watermark removal in real-world settings, outperforming previous approaches. The choice of dilation value is important to achieve the best balance between watermark removal and overall image quality.}

\section{Ablation Experiments}
In this section, we provide additional experiments related to mask generation (MG) and the inpainting process (IP). Unless otherwise stated, all \textit{IP} experiments utilize the ground truth mask for the generative process.

\subsection{[MG]: Segment Anything Model (SAM)}%
\begin{figure}[t]
    \centering
    \includegraphics[width=\columnwidth]{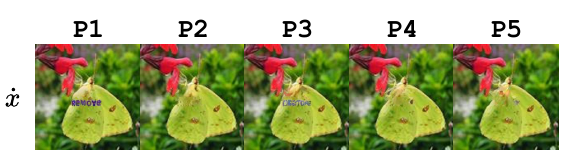}
    \caption{Visible watermark removal performance of MorphoMod using various prompts [P1-P5].}
    \label{fig:eval_prompts_small}
\end{figure}%
\begin{figure*}[tbh]
    \centering
    \includegraphics[width=0.75\textwidth]{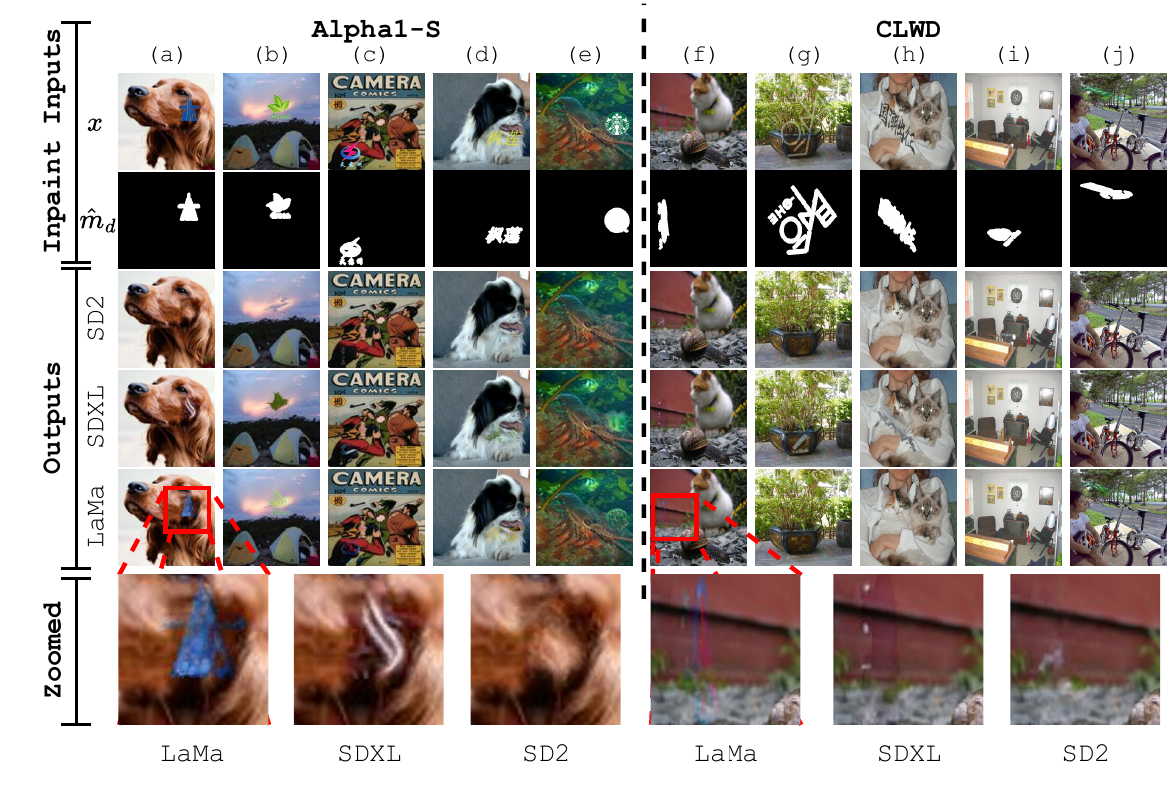}
    \caption{Visible watermark removal results with various inpainting models: LaMa, Stable Diffusion XL (SDXL), and Stable Diffusion 2 (SD2). }
    \label{fig:eval_models_img}
\end{figure*}%

A possible approach for generating watermark masks for inpainting is to use state-of-the-art segmentation models like the Segment Anything Model (SAM)~\cite{ravi2024sam2}. SAM is a prompt-driven segmentation framework trained to identify object regions using various inputs. We evaluate SAM's ability to generate watermark masks using a standalone image (\textit{Auto Mask}), an image with a ground truth bounding box (\textit{GT BBox}), an image with ground truth points (\textit{GT Points}), and an image with a generated bounding box (\textit{GDINO BBox}). The bounding box for \textit{GDINO BBox} is generated using Grounding DINO~\cite{liu2023grounding}, which takes an image and prompt to produce a bounding box. We evaluate SAM-generated masks on the Alpha1-S and CLWD datasets using F1 and IoU metrics, comparing against an SLBR baseline. All experiments are conducted with the SAM2.1 Hiera-Small model.

As shown in \tabref{tab:eval_sam}, SAM with GT BBox achieves the highest segmentation performance for Alpha1-S, but falls short of SLBR for CLWD, even with ground truth inputs (\textit{GT BBox} or \textit{GT Points}). While \textit{GT BBox} achieves the best result for Alpha1-S, this performance requires ideal conditions (ground truth vs. generated). Thus, SAM in its current state is not well-suited for watermark object detection, which the authors view positively as it limits misuse of the model. Please see the Appendix for image results.

\subsection{[IP]: Effect of Prompts in Inpainting}
\label{sec:prompts}
To evaluate how prompt choice impacts inpainting-based watermark removal, we conduct experiments using ten prompts varying in length, specificity, and style, as detailed in \tabref{tab:prompts_small}. Each prompt requests watermark removal, and we assess the output quality based on watermark removal (\(\mathrm{RMSE}_W\), \(\mathrm{SSIM}_W\), \(\mathrm{LPIPS}_W\)) and reference-free image quality metrics (BRISQUE, NIQE, PIQE) on the Alpha1-S and CLWD datasets.

Results in \tabref{tab:comb_expr} show effective watermark removal across all prompts, with high \(\mathrm{RMSE}_W\), low \(\mathrm{SSIM}_W\), and high \(\mathrm{LPIPS}_W\), alongside consistent image quality metrics. As illustrated in \figref{fig:eval_prompts_small}, the watermark is seamlessly removed in most cases. However, in some instances, inpainting introduces unintended artifacts (e.g., added text in prompts P1 and P3). Despite this, the original watermark is successfully removed, demonstrating all prompts achieve the intended goal.

\subsection{[IP]: Pre-Removal Watermark Filling}
\label{sec:prefill}
We explore the effect of pre-removal filling strategies on inpainting by testing five approaches: (1) no fill, (2) white fill, (3) black fill, (4) gray fill, and (5) average background fill, where the average color of background pixels is used. These strategies aim to assess how initial fill conditions impact inpainting results, with high-contrast fills (e.g., no fill, white, black) potentially causing abrupt transitions, while gray and average background fills may create smoother results. Watermark removal and image quality metrics (BRISQUE, NIQE, PIQE) are collected on Alpha1-S and CLWD datasets.

The results of this experiment are shown in the middle section of \tabref{tab:comb_expr}. Each fill strategy is able to successfully remove the watermark as shown by the high \(\mathrm{RMSE}_W\), low \(\mathrm{SSIM}_W\), and high \(\mathrm{LPIPS}_W\) across each dataset. For the image quality metrics, all prompts perform evenly as well. As such, the fill strategy prior to inpainting is negligible. For image results of this experiment, please see the Appendix.

\subsection{[IP]: Impact of Inpainting Model}
To evaluate the performance of different inpainting models for watermark removal, we compare three state-of-the-art open-source models: Stable Diffusion 2 (SD2), Stable Diffusion XL (SDXL), and LaMa. These models are tested on the CLWD and Alpha1-S datasets, with evaluations based on watermark removal metrics (\(\mathrm{RMSE}_W\), \(\mathrm{SSIM}_W\), \(\mathrm{LPIPS}_W\)) and image quality metrics (BRISQUE, PIQE, NIQE).

Results in \tabref{tab:comb_expr} show that all models successfully remove watermarks, as evidenced by high \(\mathrm{RMSE}_W\), low \(\mathrm{SSIM}_W\), and high \(\mathrm{LPIPS}_W\) across datasets. Image quality metrics indicate comparable performance among the models.

Visual examples in \figref{fig:eval_models_img} highlight these results. SDXL and SD2 demonstrate the best overall visual performance, but some residual traces of watermarks remain in certain images, as shown in the close-up examples. Inpainting also introduces variations during restoration, producing realistic outputs even without a target image for comparison. For example, in (a), SDXL adds subtle color enhancements, while SD2 seamlessly blends with the fur. These variations create plausible and diverse outcomes, which can be particularly valuable in real-world scenarios where exact image details are unknown.

\section{Case Study: Steganographic Disorientation}
\begin{figure}[t]
    \centering
    \includegraphics[width=0.70\linewidth]{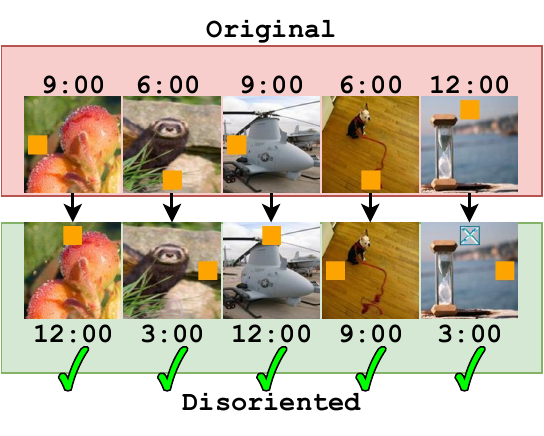}
    \caption{MorphoMod steganographic disorientation results. For each example, MorphoMod successfully disorients the original time.}
    \label{fig:steg_eval}
\end{figure}
We introduce a novel toy problem, \textit{steganographic disorientation}, to advance methods for eliminating residuals in image object removal. This problem explores disrupting information hiding by altering high-level features.

Traditional steganography hides information invisibly within low-level image features, but recent sanitization methods using diffusion models can effectively erase such messages (regeneration attacks)~\cite{robinette2023suds,robinette2024dmsuds,zhao2023invisible,Nie2022DiffusionMF}. However, when information is encoded in high-level features, like the position of objects, it persists through regeneration.

In our scheme, a message is conveyed via the position of a box in one of four predefined locations (north, east, south, or west) within an image. The goal of the disorientation agent is to alter the box's position, disrupting the intended communication (e.g., changing a meeting time). To support this, we introduce the \textit{Disorient} dataset, consisting of 4,000 training images and 1,000 test images, with orange boxes placed in four positions. A classifier and a U-Net model are trained to identify box positions and masks. The disorientation MorphoMod agent follows three steps: (1) predict the box location, (2) MorhpoMod removal of the box, and (3) add a new box in a different position from the originally predicted location. Performance is evaluated using accuracy, BRISQUE, NIQE, and PIQE metrics.
\begin{table}[t]
\caption{MorphoMod steganographic disorientation results.}
\label{tab:steg_disorient}
\vskip 0.1in
\begin{center}
\begin{small}
\begin{sc}
\begin{tabular}{c|ccc}
\toprule
    Metric & Original & Disoriented & Eval \\
    \midrule
    Accuracy & 100.0\% & 0.0\% & \textcolor{green!30}{\ding{51}} \\
    BRISQUE & 40.80 & 40.42 & \textcolor{green!30}{\ding{51}} \\
    NIQE & 4.88 & 4.85 & \textcolor{green!30}{\ding{51}} \\
    PIQE & 46.64 & 46.64 & \textcolor{green!30}{\ding{51}} \\
 
\bottomrule
\end{tabular}
\end{sc}
\end{small}
\end{center}
\vskip -0.1in
\end{table}

The results of this experiment are shown in \tabref{tab:steg_disorient}. From these results our inpainting disorientation model is successful at disorienting the communication as highlighted by the 0.0\% accuracy. The image results for this experiment are shown in \figref{fig:steg_eval}. In each example, the original time is successfully shifted to a new time. 

While we present this toy problem to the field, we would like to highlight challenges for future consideration. How to ensure that the disorientation goes unnoticed?  If residual artifacts remain in the image, the recipient can detect the tampering and infer that the message has been altered.

\section{Discussion and Conclusion}
We introduce MorphoMod, an automated method for visible watermark removal that operates without requiring target images, making it highly effective in real-world scenarios. Extensive experiments on traditional and newly proposed Alpha1 datasets demonstrate MorphoMod’s high performance in watermark removal while maintaining semantic quality. Our results highlight its adaptability across varying configurations, with ablation studies showcasing the impact of prompts, inpainting strategies, and pre-removal filling methods. Additionally, a novel case study on steganographic disorientation demonstrates MorphoMod’s potential for broader applications, such as disrupting hidden information in images. In the future, we would like to explore additional methods for generative watermark localization. \textit{MorphoMod sets a new benchmark for blind watermark removal, paving the way for future advancements in watermarking and image restoration research.}

\section{Impact Statement}
MorphoMod introduces a novel approach to blind visible watermark removal, eliminating the need for target images. By leveraging morphological dilation and generative inpainting, MorphoMod effectively removes opaque and transparent watermarks while preserving image integrity. Our work significantly enhances real-world applicability, as demonstrated by its high performance on traditional and newly proposed benchmark datasets.

As such, we do not take the implications of this work lightly. \textbf{We believe this work provides the necessary empirical evidence needed to assess current generative watermark removal performance and to urge the development of more resilient digital watermarking techniques that are suited for the current generative landscape.}






\bibliography{references}
\bibliographystyle{icml2025}

\newpage
\appendix
\onecolumn
\section{Appendix}
\subsection{Dataset Details}
\begin{enumerate}
    \item \textbf{CLWD:} CLWD consists of 60K watermarked images for training and 10K for testing. The watermarks are gathered from publicly available logo websites and applied to images randomly selected from PASCAL VOC2012~\cite{everingham2015pascal}. The transparency ranges from 0.3 to 0.7, and its size, position, orientation, and transparency are all randomly selected across individual images.\\
    \item \textbf{LOGO-Gray:} LOGO-Gray contains 12K training and 2K testing samples with gray-scale watermarks. The size and transparency of the watermark are between 35\% and 85\%.  \\
    \item \textbf{LOGO-L:} LOGO-L contains 12K training and 2K testing samples. The transparency of the watermarks ranges from 35\% to 60\%, and the size is anywhere from 35\% to 60\% of the width of the host images. \\
    \item \textbf{LOGO-H:} LOGO-H is a harder subset of LOGO-L and contains 12K training and 2K testing samples. The transparency and size are randomly chosen from 60\% to 85\%.
\end{enumerate}

\subsection{Additional Experiments and Results}
\subsubsection{[IP]: Effect of Prompts in Inpainting Cont.}
This section provides the full table of prompts and results for \secref{sec:prompts}.

\begin{table}[h]
\caption{Prompts used for inpainting.}
\label{tab:prompts}
\vskip 0.15in
\begin{center}
\begin{small}
\begin{tabular}{c|l}
    \toprule
        \textsc{ID} & \textsc{ Prompt}  \\
        \midrule
        \textsc{P1} & Remove. \\
        \textsc{P2} & Fill in the background. \\
        \textsc{P3} & Erase the mark and restore the original. \\
        \textsc{P4} & Blend into the surrounding area. \\
        \textsc{P5} & Reconstruct the missing details. \\
        \textsc{P6} & Remove the object and match the background. \\
        \textsc{P7} & Fill in the gaps as if the mark was never there. \\
        \textsc{P8} & Smooth out and complete the scene. \\
        \textsc{P9} & Mend the area to look natural. \\
        \textsc{P10} &  Restore the natural texture. \\
    \bottomrule
\end{tabular}
\end{small}
\end{center}
\vskip -0.1in
\end{table} 

\begin{table*}[h]
\caption{Inpainting experimental results of various experiments on the Alpha1-S and CLWD datasets for $RMSE_W$ ($R_W$), $SSIM_W$ ($S_W$), $LPIPS_W$ ($LP_W$), BRISQUE (B), NIQE (N), and PIQE (P).  }
\label{tab:prompt_results_full}
\vskip 0.1in
\begin{center}
\begin{small}
\begin{sc}
\begin{tabular}{c|cccccc|cccccc}
\toprule
 \multirow{2}{*}{Prompt} &  \multicolumn{6}{c|}{Alpha1-S} & \multicolumn{6}{c}{CLWD} \\ 
  & $R_W (\uparrow)$ & $S_W (\downarrow)$ & $LP_W (\uparrow)$ & B & N & P & $R_W (\uparrow)$ & $S_W (\downarrow)$ & $LP_W (\uparrow)$ & B $(\downarrow)$ & $(\downarrow)$ & P  $(\downarrow)$\\
 \midrule
P1 & 0.369 & 0.599 & 0.317 &  38.88 & 4.96 & 43.49 & 0.216 & 0.791 & 0.176 &  25.24 & 4.39 & 38.49 \\
 P2 & 0.369 & 0.599 & 0.318 &  39.34 & 4.97 & 43.29 & 0.211 & 0.795 & 0.173 &  25.41 & 4.37 & 38.47 \\
 P3 & 0.373 & 0.598 & 0.317 &  38.96 & 4.96 & 43.23 & 0.223 & 0.786 & 0.181 &  24.78 & 4.38 & 38.45 \\
 P4 & 0.370 & 0.599 & 0.317 &  39.21 & 4.95 & 43.34 & 0.214 & 0.793 & 0.174 &  25.45 & 4.34 & 38.41 \\
 P5 & 0.371 & 0.598 & 0.317 &  39.18 & 4.99 & 43.33 & 0.216 & 0.792 & 0.176 &  24.86 & 4.36 & 38.28 \\
 P6 & 0.377 & 0.592 & 0.321 &  39.20 & 4.98 & 43.39 & 0.216 & 0.792 & 0.178 &  25.22 & 4.35 & 38.72 \\
 P7 & 0.375 & 0.594 & 0.323 &  39.07 & 4.99 & 43.47 & 0.216 & 0.792 & 0.176 &  25.27 & 4.35 & 38.55 \\
 P8 & 0.372 & 0.595 & 0.320 &  39.31 & 4.95 & 43.44 & 0.214 & 0.794 & 0.175 &  25.26 & 4.32 & 38.43 \\
 P9 & 0.373 & 0.596 & 0.318 &  39.05 & 4.97 & 43.55 & 0.217 & 0.791 & 0.177 &  25.99 & 4.37 & 38.77 \\
 P10 & 0.369 & 0.599 & 0.317 &  39.12 & 4.94 & 43.26 & 0.214 & 0.794 & 0.174 &  25.97 & 4.30 & 38.57 \\
 
\bottomrule
\end{tabular}
\end{sc}
\end{small}
\end{center}
\vskip -0.1in
\end{table*} 

\begin{figure*}[t]
    \centering
    \includegraphics[width=0.90\textwidth]{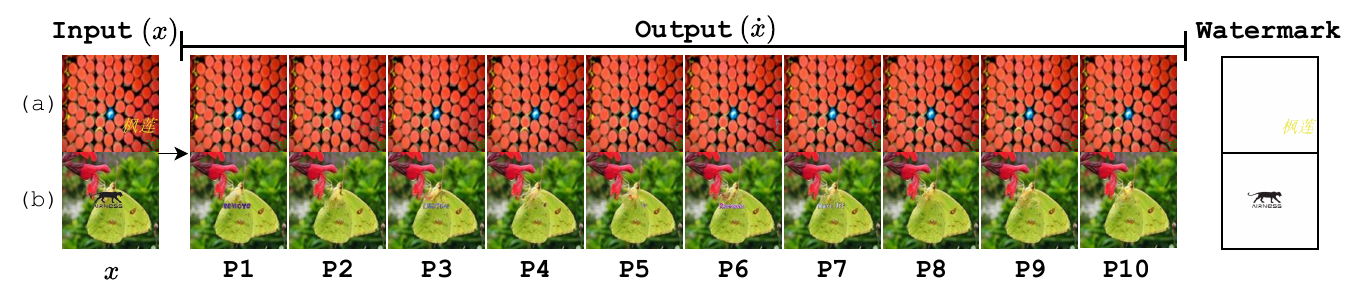}
    \caption{Visible watermark removal performance of MorphoMod using various prompts [P1-P10]. In some instances, inpainting introduces unintended artifacts (e.g., added text in prompts P1, P3, P6, and P7). Despite this, the original watermark is successfully removed, demonstrating all prompts achieve the intended goal. }
    \label{fig:eval_prompts_img}
\end{figure*}

\subsubsection{[IP]: Effect of Denoising Steps}
\begin{figure*}[ht]
\vskip 0.2in
\begin{center}
\begin{subfigure}[t]{0.45\textwidth}
    \centerline{\includegraphics[width=\columnwidth]{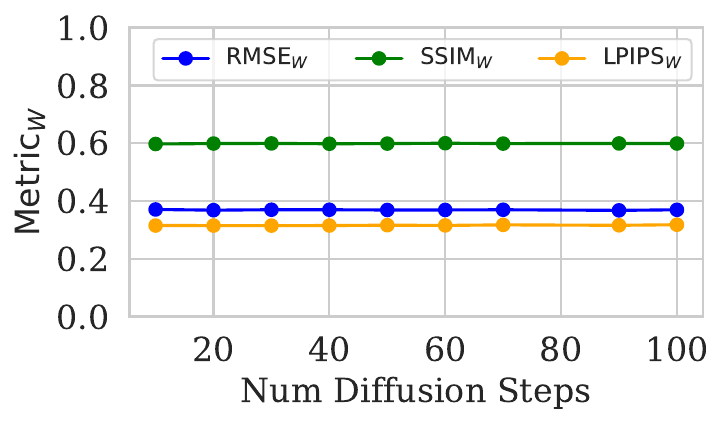}}
    \caption{Watermark Metrics}
    \label{fig:eval_steps_w}
\end{subfigure}
\hfill
\begin{subfigure}[t]{0.45\textwidth}
    \centerline{\includegraphics[width=\columnwidth]{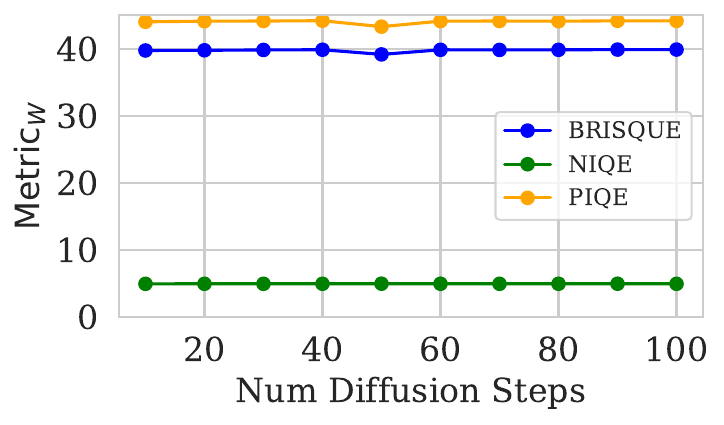}}
    \caption{Image Quality Metrics}
    \label{fig:eval_steps_im}
\end{subfigure}
\caption{Watermark and image quality metrics for different diffusion steps in the inpainting process for Stable Diffusion 2. }
\label{fig:eval_steps}
\end{center}
\vskip -0.2in
\end{figure*}

In the inpainting process, the number of denoising steps is associated with the iterative refinement of the image construction. To investigate the impact of different step counts, we conduct experiments using $s \in \{10, 20, 30, 40, 50, 60, 70, 80, 90, 100\}$ denoising steps. For each configuration, we evaluate using the Alpha1-S and CLWD datasets and collect watermark removal and BRISQUE, PIQE, and NIQE image quality metrics. From the results shown in \figref{fig:eval_steps}, the number of diffusion steps has no impact on watermark removal performance. 

\begin{figure*}[t]
    \centering
    \includegraphics[width=0.80\textwidth]{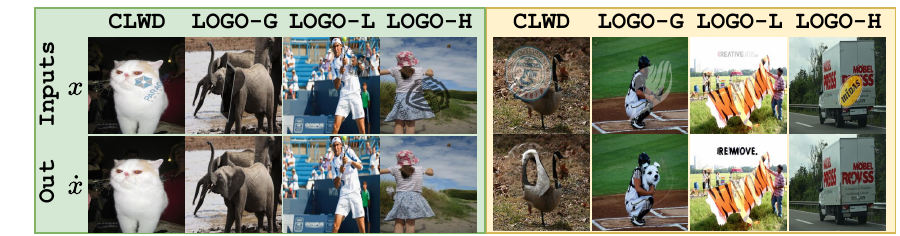}
    \caption{\textbf{[Experiment 1]} Visible watermark removal results of MorphoMod on various datasets. The left-hand side of the image demonstrates removal that is imperceptible, and the right-hand side of the image demonstrates removal that \textit{is} perceptible. For instance, the dog being generated on the baseball field (\textit{LOGO-G}). A human would be able to quickly notice that something is not right in this image. These results highlight an important area for future work: how to ensure the generated areas are imperceptible to the human eye? What metrics could be used to evaluate this?}
    \label{fig:dilation_ex}
\end{figure*}

\begin{figure*}[t]
    \centering
    \includegraphics[width=\textwidth]{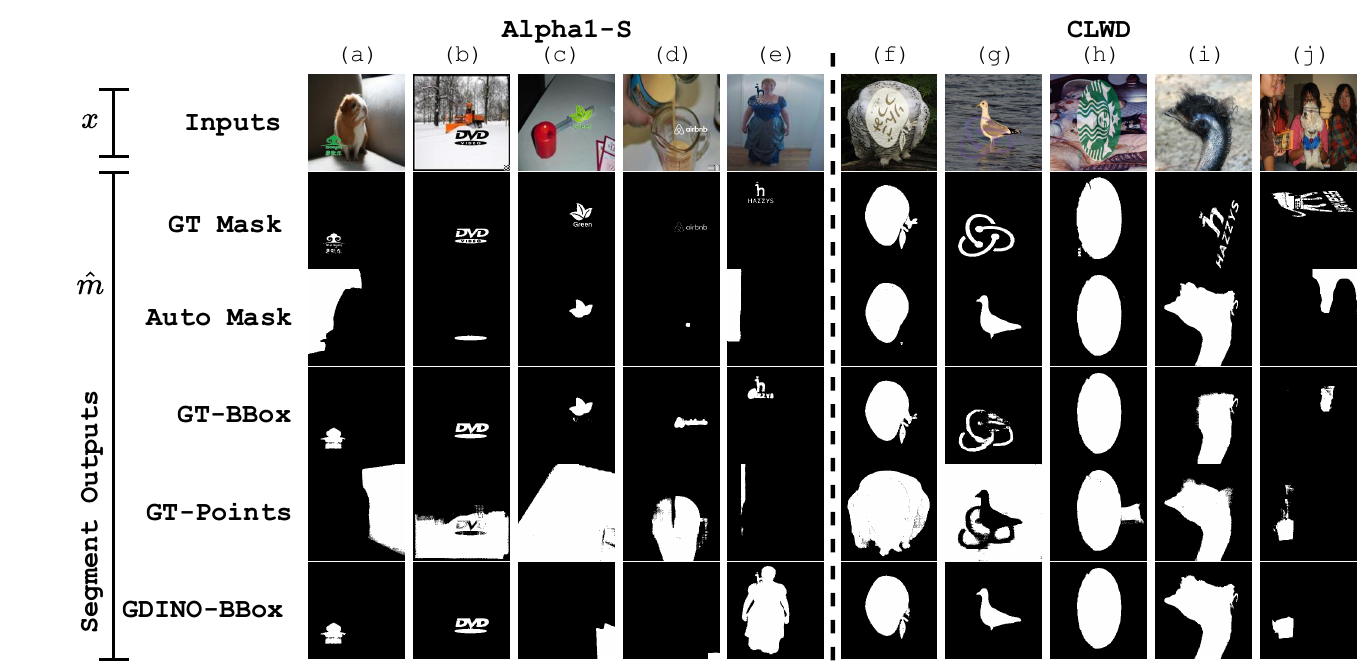}
    \caption{\textbf{[5.1 Results]} Segmentation masks generated with different SAM strategies including (1) using a standalone image (\textit{Auto Mask}), (2) an image with a ground truth bounding box (\textit{GT BBox}), (3) an image with ground truth points (\textit{GT Points}), and (4) an image with a generated bounding box (\textit{GDINO BBox}). The bounding box for \textit{GDINO BBox} is generated using Grounding DINO~\cite{liu2023grounding}, which takes an image and prompt to produce a bounding box.}
    \label{fig:eval_sam}
\end{figure*}
\begin{figure}[!ht]
    \centering
    \includegraphics[width=0.60\linewidth]{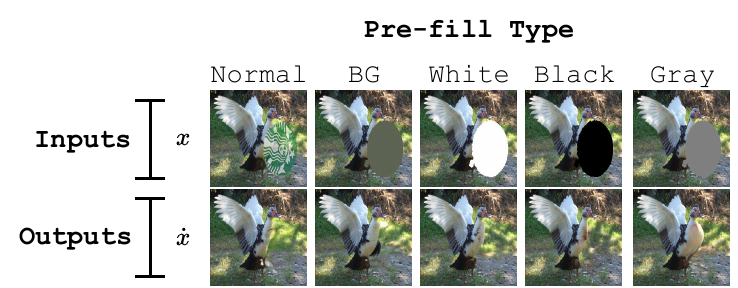}
    \caption{\textbf{[5.3 Results]} Pre-fill examples including (1) normal, (2) average background [BG], (3) white, (4) black, and (5) gray strategies. $\restoreout$ is the resulting restored image from each watermarked image $\inputim$.}
    \label{fig:fill_ex}
\end{figure}
\begin{figure*}[t]
    \centering
    \includegraphics[width=\textwidth]{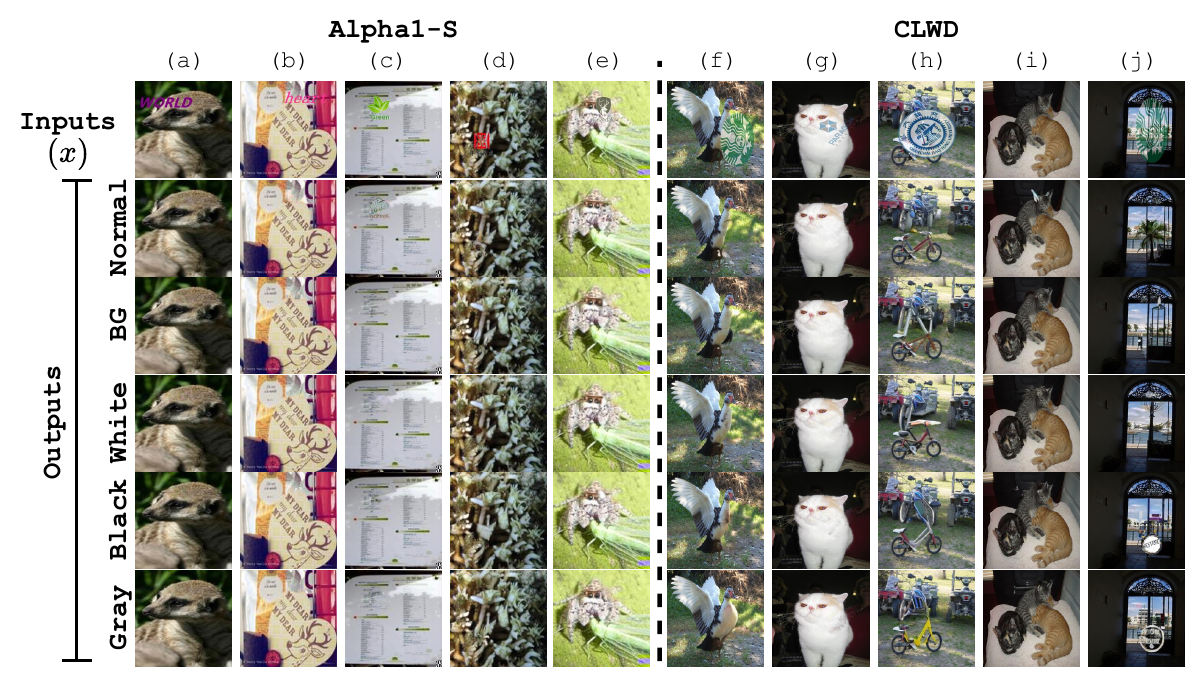}
    \caption{\textbf{[5.3 Results]} Example images generated from different fill strategies as described in \secref{sec:prefill}.}
    \label{fig:eval_fill_img}
\end{figure*}


\end{document}